\documentclass[a4paper]{article}

\usepackage[utf8]{inputenc}
\usepackage{erk}
\usepackage{times}
\usepackage{graphicx}
\usepackage{placeins}
\usepackage{caption}
\usepackage[top=22.5mm, bottom=22.5mm, left=22.5mm, right=22.5mm]{geometry}
\usepackage{url}
\usepackage[slovene,english]{babel}

\def\footnotemark{}

\begin{document}
\title{Evaluation of Image Matching Methods for Visual Odometry on UAVs}

\author{Gašper Spagnolo$^{1}$, Luka Čehovin Zajc$^{1}$, Matej Dobrevski$^{1}$} 

\affiliation{$^{1}$Faculty of Computer and Information Science, University of Ljubljana,\\
Večna pot 113, Ljubljana, Slovenia.}

\email{E-pošta: matej.dobrevski@fri.uni-lj.si}

\maketitle

\sloppy

\begin{abstract}{Abstract}
Unmanned aerial vehicles (UAVs) are becoming a powerful tool for many environmental monitoring and transport applications. Yet, their reliance on Global Navigation Satellite System (GNSS) technology for navigation makes them susceptible to catastrophic failures in scenarios where the positioning signal is unavailable or disrupted. This work explores Visual Odometry (VO) as a crucial navigation component. Recently, numerous deep-learning-based methods for image matching have been proposed that are yet to be implemented in a fully-fledged VO system. In this paper, we evaluate recent state-of-the-art image matching methods for the task of VO for UAV position tracking, with a downwards-facing camera, on our synthetic dataset, and find that while the best results are generated by the recent RoMa matcher, SIFT features can outperform some recent state-of-the-art.

\end{abstract}

\selectlanguage{english}

\section{Introduction}

Unmanned Aerial Vehicles (UAVs) most commonly rely on GNSS technology to determine their location in the environment. However, GNSS technology is not always reliable; it is vulnerable to jamming, spoofing, and environmental interferences obstructing radio signals. Looking even further, GNSS is unavailable for extraterrestrial missions, such as landing and navigating on Mars~\cite{mars}. \par

\begin{figure}
\centering
\includegraphics[width=\linewidth]{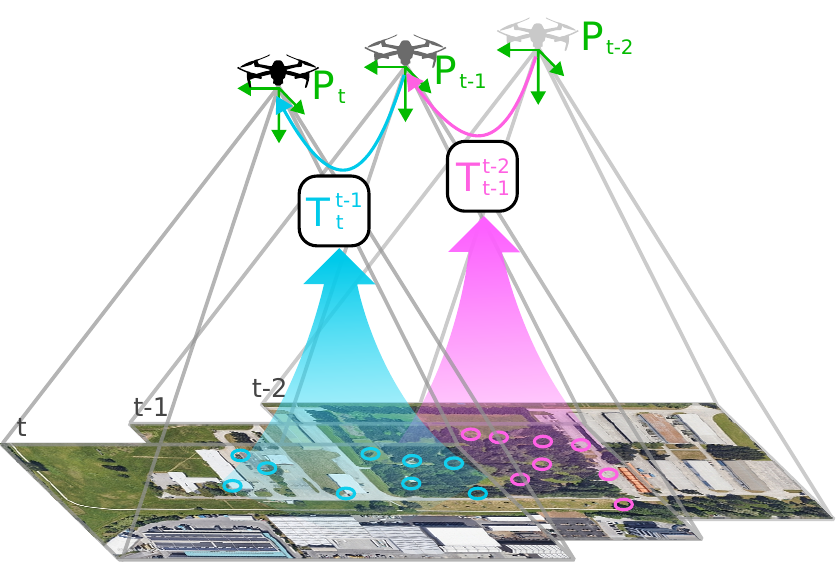}
\caption{Visual odometry for UAV position estimation. Overlapping images from the drone are matched, and a relative camera transformation is estimated based on matches. Finally, local transformations between images are aggregated.}
\label{fig:overview}
\end{figure}

An alternative approach to localization is the use of cameras. Visual Odometry (VO) aims to estimate the position and orientation of a moving camera by analyzing the changes in images captured over time. As illustrated in Figure~\ref{fig:overview}, it relies on image-matching techniques to find pixel correspondences and uses them to estimate a camera transformation. The core of most Visual Odometry approaches is an image-matching method. For a long time, image matching was mostly done by using hand-crafted keypoint detectors and descriptors~\cite{SIFT, SURF, BRIEF, ORB} and descriptor distance metrics for matching, such as Mutual Nearest Neighbours (MNN). Recently, deep-learning-based approaches for keypoint detection and description (keypoint extraction)~\cite{superpoint, ALIKED, SILK, dedode}, combined with deep-learning-based keypoint matching~\cite{superglue, lightglue} have started to outperform classical approaches on standard benchmarks like KITTI Visual Odometry Benchmark~\cite{KITTI}. Furthermore, approaches have appeared that perform image matching end-to-end without separating the process into keypoint detection, description, and matching, treating all the pixels as keypoints.\par

Images captured from a UAV with a down\-ward-facing camera have a significantly different distribution from those in most VO benchmarks, where the camera is usually mounted on a ground vehicle or handheld horizontally facing. Namely, the camera is more likely to rotate about its Z-axis, and the key points will likely be on the same plane in space. Both of these are known degenerate cases for calculating the essential matrix from pixel correspondences. In this paper, we evaluate recent state-of-the-art image-matching methods to the problem of estimating the relative camera position for a downward-facing camera mounted on a UAV (Figure~\ref{fig:overview}). For this purpose, we acquired a synthetic dataset of three flyovers over different central European cities. We evaluate the keypoint matching, the estimated rotations and translations, and the final estimated trajectory.

\section{Related work}

For a long time, approaches to this problem have been based on hand-engineered methods~\cite{SIFT,SURF,BRIEF,ORB} for local feature (keypoint) detection and description, using some distance metrics in feature space, like Mutual Nearest Neighbor (MNN) or Lowe's ratio test, to match the keypoints across images. Deep-learning-based keypoint extractors have recently started to outperform the traditional methods, e.g., SuperPoint~\cite{superpoint}, DeDoDe~\cite{dedode}, SiLK~\cite{SILK}, and ALIKED~\cite{ALIKED}. Deep methods for keypoint matching, like SuperGlue~\cite{superglue} and LightGlue~\cite{lightglue} have also started to outperform simpler techniques like MNN for feature matching on public benchmarks~\cite{hpatches_2017_cvpr, inloc}\par

Dense matchers, like LoFTR~\cite{LoFTR}, ASpanFormer~\cite{ASpanFormer}, CasMTR~\cite{CasMTR}, RoMa~\cite{roma}, and others, take a different approach to image matching and replace the keypoint detection with dense matching, treating every pixel as a key point. Dense approaches estimate a dense warp, with the goal of estimating every matchable pixel pair.\par

DUSt3R~\cite{dust3r} is a recent fundamentally different approach that performs image matching in 3D space. It performs (normalized) depth estimation for given input image pairs and outputs align\-ed point maps (point clouds with pixel correspondences). Matched points are extracted from the aligned point clouds as the points with the smallest 3D distance. MASt3R~\cite{mast3r}, extends DUSt3R and is explicitly trained for point matching. It outputs additional per-pixel features that are optimized for matching.\par


\section{Methods and Datasets}

In this paper, we evaluate five different image-matching methods. We have primarily focused on recent deep-learning-based approaches and included a well-perfor\-ming hand-crafted method.

\begin{itemize}

\item \textbf{SIFT}~\cite{SIFT} is a popular and time-tested hand-crafted method for detecting and describing local features (keypoints) in images, invariant to scaling and rotation. In this work we used Lowe's ratio test (LRT) for matching SIFT keypoints.


\item \textbf{SuperGlue} (SG)~\cite{superglue} uses a deep Graph Convolution Network (GNN) for keypoint matching. SuperGlue must be trained for the specific keypoint extraction method it is used with. In this paper, we use SuperGlue with keypoints extracted by SuperPoint (SP).

\item \textbf{LightGlue} (LG)~\cite{lightglue} is a transformer-based keypoint mat\-ching network whose depth adapts at test time to trade speed and accuracy. In this paper, we use LightGlue with keypoints extracted by SuperPoint.

\item \textbf{Robust Dense Feature Matching} (RoMa)~\cite{roma} is a deep dense image matcher that treats all the pixels as keypoints, estimating a wrap based on their description in feature space. Recently, it demonstrated for the first time that dense matchers can outperform sparse matchers, on standard benchmarks.\par

\item \textbf{MASt3R}~\cite{mast3r} is a recent method that extends~\cite{dust3r} to specifically address the problem of pixel matching. The model predicts two aligned pointmaps (pointcloud with pixel correspondences) for the two input images. It also outputs descriptions for each pixel/3D point and contains an output head that performs an iterative procedure for matching the descriptors.\par
\end{itemize}

\begin{figure}[!t]
\centering
\includegraphics[width=\linewidth]{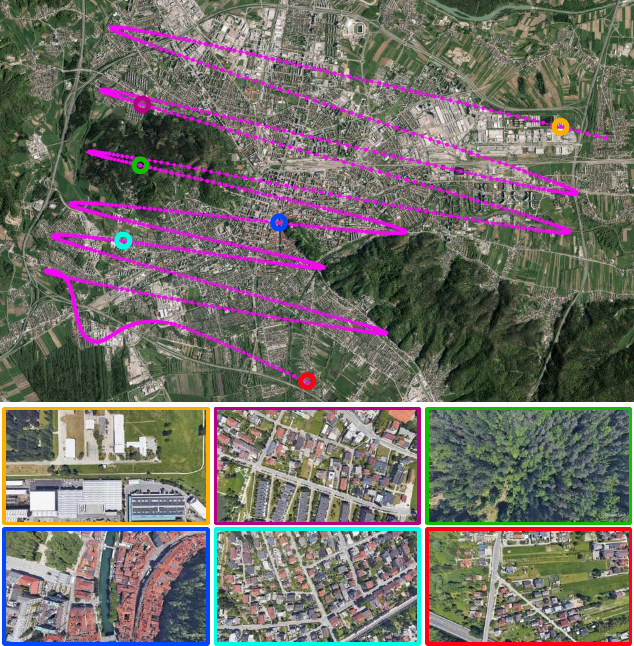}
\caption{Overview of Ljubljana sequence in the dataset. The sequences contain images of diverse urban, suburban, agricultural, and forestry scenery.}
\label{fig:dataset}
\end{figure}

Our dataset for evaluation consists of three sequences, each containing $3000$ images. The images are generated using Google Earth Studio\footnote{\url{https://www.google.com/earth/studio/}} (GES). The scenes are Ljubljana, Graz, and Venice. The camera was set at $150~m$ above each city's reported above-sea-level height in the dataset, occasionally elevating up to $850~m$. The camera was set to always point directly downwards, emulating an inertially stabilized camera. The camera properties were set to $80^{\circ}$ vertical FOV and a resolution of $1920\times1080$. The camera trajectory was manually drawn to cover a large part of the city, going over urban areas, suburban areas, and areas covered with vegetation. The camera trajectory for the Ljubljana sequence and examples of images are shown in Figure~\ref{fig:dataset}. The dataset will be publicly released in the coming months. \par

\subsection{Visual Odometry}

The keypoint extraction approaches were integrated into a simple VO pipeline. Keypoints were extracted and mat\-ched in successive images, skipping over images with relative camera translation less than $70~m$ (using the ground truth annotations). Then, the matched keypoints were used to fit both an essential matrix and a homography, using OpenCV's implementation of RANSAC, with an inlier threshold set to $1px$ and set to perform $10000$ iterations. Then, the relative rotation and translation were extracted, using a chirality check to select the physically plausible solution. The translation vector was then scaled to the ground truth distance between the camera frames. Finally, the complete path was reconstructed in the first camera frame of the first image.


\section{Results}
\subsection{Keypoint extraction and matching}
Most methods had no trouble extracting and matching keypoints on most frames, as can be seen in Table~\ref{tab_keys}. Note that for RoMa the number of matched keypoints is a parameter that we set.

\setlength{\tabcolsep}{4pt}
\begin{table}[h]
\caption{Number of matched keypoints.} \label{tab_keys}
\vspace{-0.5cm}
\begin{center}
\begin{tabular}{ r | c c | c c | c c}
 & \multicolumn{2}{c|}{Ljubljana} & \multicolumn{2}{c|}{Graz} & \multicolumn{2}{c}{Venice}\\
\hline  
  \textbf{Method}& \textbf{Min.}  & \textbf{Max.} & \textbf{Min.}  & \textbf{Max.}&\textbf{Min.} & \textbf{Max.} \\ 
\hline  
  SIFT+LRT &  889 & 24K & 1186 & 25K & 88 & 32K\\
  SP+SG & 8 & 1980 & 2 & 1825 & 45 & 1866\\
  SP+LG & 0 & 5641 & 1& 4738 & 124 & 4935\\
  RoMa & 10K & 10K& 10K & 10K & 10K & 10K\\
  MASt3R & 1257 & 2304 & 1300& 2304 & 885 & 2278\\
\hline  
\end{tabular}
\end{center}
\end{table}
\vspace{-0.5cm}
\setlength{\tabcolsep}{4pt}
\begin{table}[h]
\caption{Failures, as the number of frames where a method found less than 8 or 100 matches.} \label{tab_fails}
\vspace{-0.5cm}
\begin{center}
\begin{tabular}{ r | c c | c c | c c}
 & \multicolumn{2}{c|}{Ljubljana} & \multicolumn{2}{c|}{Graz} & \multicolumn{2}{c}{Venice}\\
\hline  
  \textbf{Method}& \textbf{$<8$}  & \textbf{$<100$} & \textbf{$<8$}  & \textbf{$<100$}&\textbf{$<8$} & \textbf{$<100$} \\ 
\hline  
  SIFT+LRT &  0 & 0 & 0 & 0 & 0 & 1\\
  SP+SG & 0 & 2 & 2 & 3 & 0 & 24\\
  SP+LG & 2 & 2 & 3 & 3 & 0 & 0\\
  RoMa & 0 & 0 & 0 & 0 & 0 & 0\\
  MASt3R & 0 & 0 & 0 & 0 & 0 & 0\\
\hline  
\end{tabular}
\end{center}
\end{table}
\vspace{-0.5cm}

\begin{figure*}[!t]
{\scriptsize SIFT+LRT \hspace{2.1cm}  SP+SG \hspace{2.5cm}  SP+LG  \hspace{2.5cm}  RoMa \hspace{2.55cm}  MASt3R}\\
  \includegraphics[width=\textwidth]{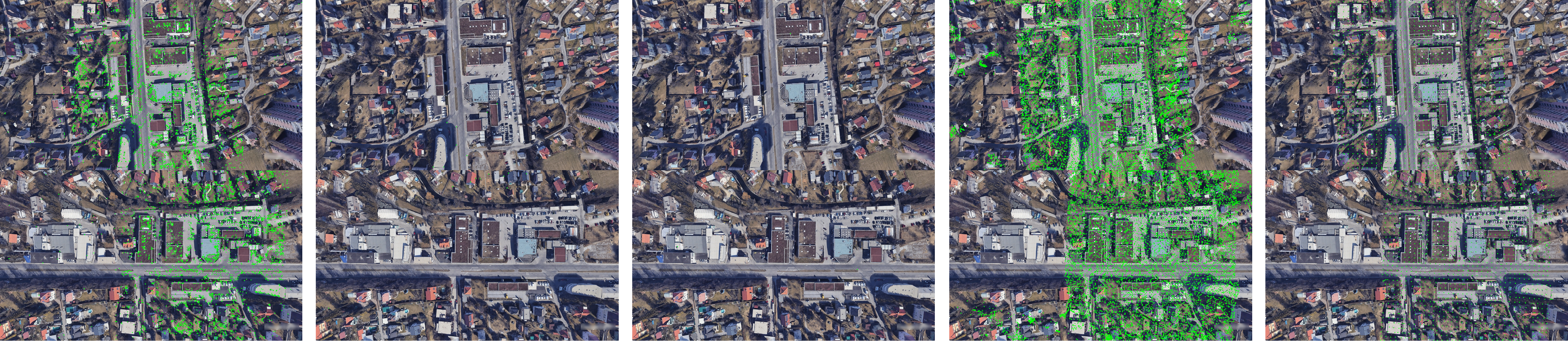}
  \caption{A comparison of the matched keypoints by different methods on sequential frames with large rotation from the Graz sequence. SIFT+LRT, RoMa and MASt3R have no problems finding matches. Since RoMa and MASt3R match all (matchable) pixels, the keypoints cover a rectangular region. SuperPoint, SuperGlue and LightGlue are sensitive to rotation, and produce relatively few matches.} \label{fig:fails_graz}
\end{figure*}

For estimating the essential matrix or the homography, less than $8$ or $4$ keypoints is an absolute failure. Finding less than $\sim 100$ matches is likely also a failure, as there is a high likelihood that the fitted transformations will not be very good. As can be seen in Table~\ref{tab_fails}, most methods suffered no absolute failures, except for SuperPoint with SuperGlue and LightGlue. We found that these failures were due to large rotations about the camera $Z$ axis. An example containing an extreme rotation is shown in Figure~\ref{fig:fails_graz}.

\subsection{Essential matrix and homography estimation}

The depth of the 3D points in our dataset, for the most part, is significantly smaller than the depth from the camera to the scene, rendering them approximately planar. This is a degenerate case for the estimation of the essential matrix, as it loses rank.\par

\setlength{\tabcolsep}{4pt}
\begin{table}[!h]
\caption{Average absolute error in the estimated direction of the translation vector [degrees].} \label{tab_dir}
\vspace{-0.5cm}
\begin{center}
\begin{tabular}{ r | c c | c c | c c}
 & \multicolumn{2}{c|}{Ljubljana} & \multicolumn{2}{c|}{Graz} & \multicolumn{2}{c}{Venice}\\
\hline  
  \textbf{Method}& \textbf{Ess.}  & \textbf{Hom.} & \textbf{Ess.}  & \textbf{Hom.}&\textbf{Ess.} & \textbf{Hom.} \\ 
\hline  
  SIFT+LRT &  0.01 & 0.01 & 0.10 & \textbf{0.09} & 0.03 & 0.03\\
  SP+SG & \textbf{0.24} & 0.33 & \textbf{0.20} & 0.49 & 0.22 & \textbf{0.07}\\
  SP+LG & 0.04 & 0.04 & \textbf{0.11} & 0.36 & 0.10 & \textbf{0.06}\\
  RoMa & 0.01 & 0.01 & 0.10 & \textbf{0.09} & 0.02 & \textbf{0.01}\\
  MASt3R & 0.09 & \textbf{0.02} & 0.20 & \textbf{0.10} & 0.14 & \textbf{0.03}\\
\hline  
\end{tabular}
\end{center}
\end{table}

For this reason, we also fit a homography to the matched keypoints, hypothesizing that most of the time there will be enough points lying on the same plane in 3D. Our hypotesis was confirmed by the results in Tables~\ref{tab_dir} and \ref{tab_orient}. The rotations and translations extracted from the homography, in general, are more accurate than the ones extracted from the essential matrix for our dataset.

\begin{table}[h]
\caption{Average absolute error in the estimated rotation about the camera Z axis [degrees].} \label{tab_orient}
\vspace{-0.5cm}
\begin{center}
\begin{tabular}{ r | c c | c c | c c}
 & \multicolumn{2}{c|}{Ljubljana} & \multicolumn{2}{c|}{Graz} & \multicolumn{2}{c}{Venice}\\
\hline  
  \textbf{Method}& \textbf{Ess.}  & \textbf{Hom.} & \textbf{Ess.}  & \textbf{Hom.}&\textbf{Ess.} & \textbf{Hom.} \\ 
\hline  
  SIFT+LRT &  \textbf{0.03} & 0.05 & 0.03 & 0.03 & 0.09 & \textbf{0.04}\\
  SP+SG & 0.71 & \textbf{0.25} & 0.66 & \textbf{0.24} & 1.26 & \textbf{0.10}\\
  SP+LG & 0.22 & \textbf{0.12} & 0.15 & \textbf{0.11} & 0.64 & \textbf{0.10}\\
  RoMa & \textbf{0.03} & 0.04 & 0.06 & \textbf{0.03} & 0.07 & \textbf{0.03}\\
  MASt3R & 1.04 & \textbf{0.11} & 0.77 & \textbf{0.06} & 1.16 & \textbf{0.07}\\
\hline  
\end{tabular}
\end{center}
\end{table}

\vspace{-0.5cm}

\begin{figure}
\centering
{\scriptsize SIFT+LRT \hspace{0.7cm}  SP+SG \hspace{0.7cm}  SP+LG  \hspace{0.9cm}  RoMa \hspace{0.9cm}  MASt3R}
\includegraphics[width=\linewidth]{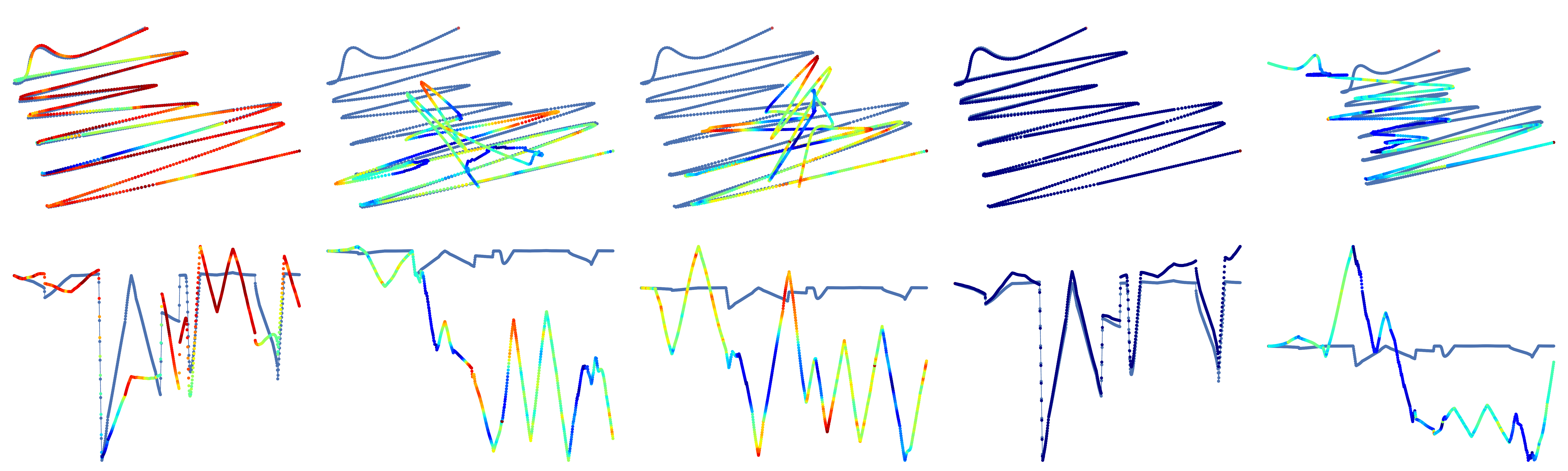}
\medskip {\scriptsize (a) Ljubljana: Total length $84.3~km$, above-sea altitude from $440-1170~m$.}
\includegraphics[width=\linewidth]{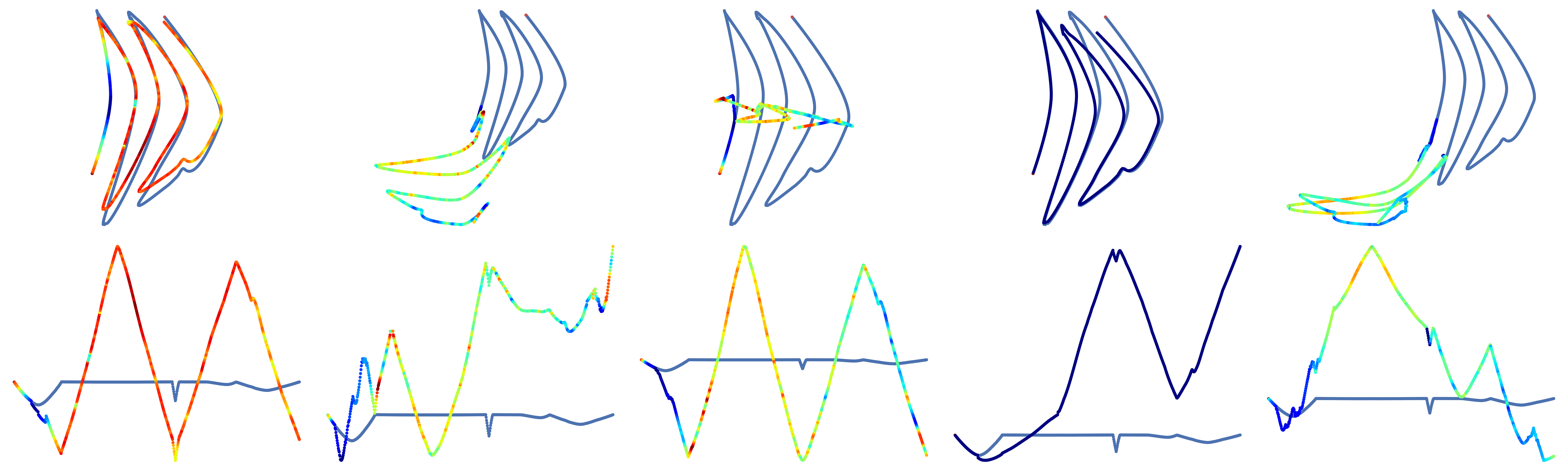}
\medskip {\scriptsize (a) Graz: Total length $57.3~km$, above-sea altitude from $503-1053~m$.}
\includegraphics[width=\linewidth]{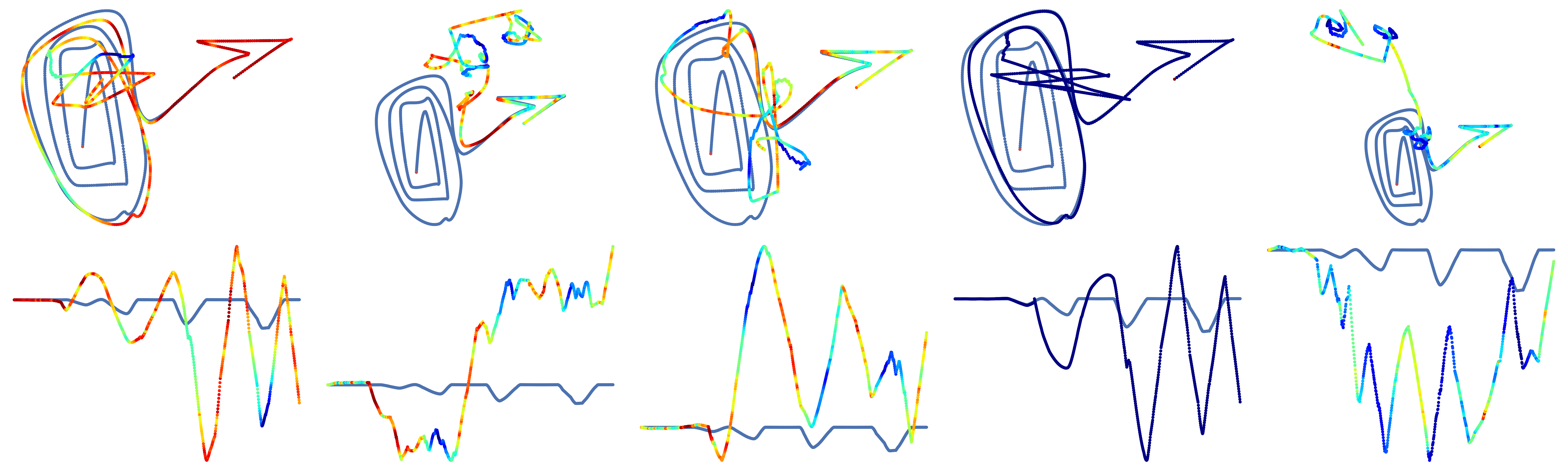}
\medskip {\scriptsize (a) Venice: Total length $81.2~km$, above-sea altitude from $144-992~m$.}
\caption{Visualizations of the estimated trajectories, using the estimated \textbf{essential matrix} to calculate the rotation and the translation. Red - small number of matched keypoints, blue - large number (normalized per sequence).}
\label{fig:ess_traj}
\end{figure}

\FloatBarrier
\subsection{Visual odometry}

\begin{figure}
\centering
{\scriptsize SIFT+LRT \hspace{0.7cm}  SP+SG \hspace{0.7cm}  SP+LG  \hspace{0.9cm}  RoMa \hspace{0.9cm}  MASt3R}
\includegraphics[width=\linewidth]{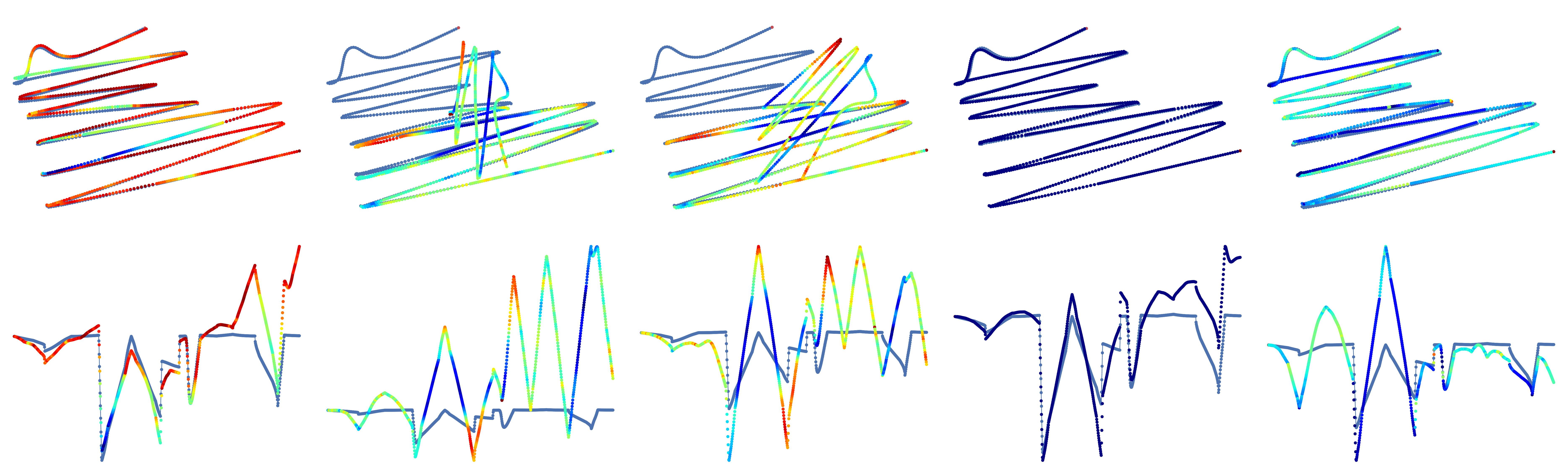}
\medskip {\scriptsize (a) Ljubljana: Total length $84.3~km$, above-sea altitude from $440-1170~m$.}
\includegraphics[width=\linewidth]{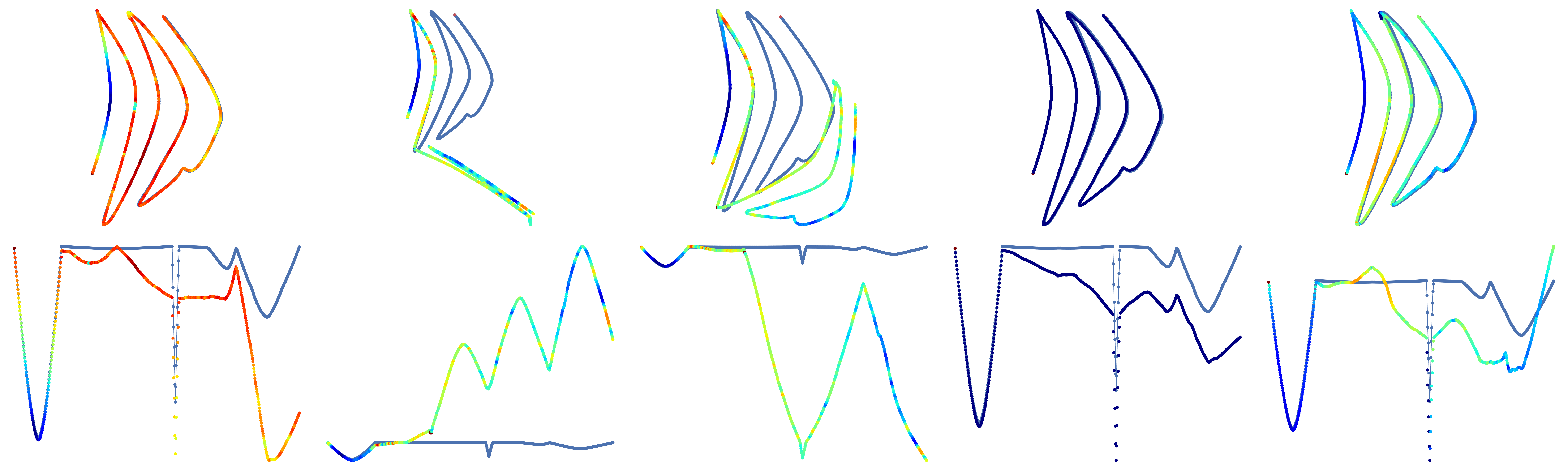}
\medskip {\scriptsize (a) Graz: Total length $57.3~km$, above-sea altitude from $503-1053~m$.}
\includegraphics[width=\linewidth]{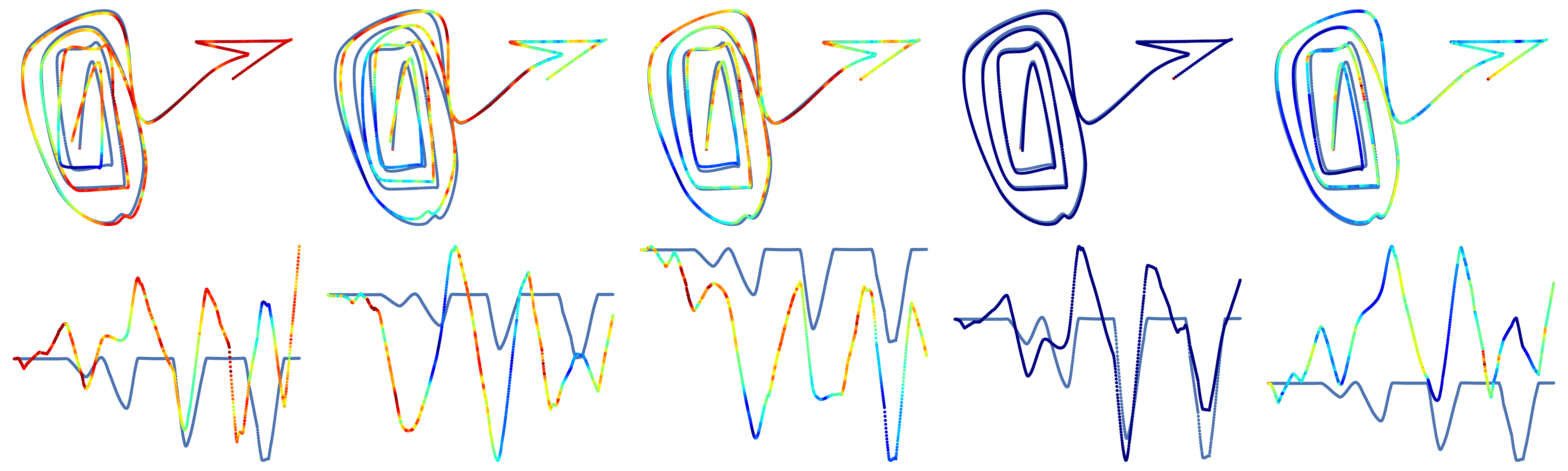}
\medskip {\scriptsize (a) Venice: Total length $81.2~km$, above-sea altitude from $144-992~m$.}
\caption{Visualizations of the estimated trajectories, using the estimated \textbf{homography} to calculate the rotation and the translation. Red - small number of matched keypoints, blue - large number (normalized per sequence).}
\label{fig:homo_traj}
\end{figure}

The relative translations and rotations were chained together to estimate the complete trajectory of the camera. The results obtained using the essential matrix are shown in Figure~\ref{fig:ess_traj} and the results obtained using the homography are shown in Figure~\ref{fig:homo_traj}. We can see that the trajectories obtained through the homography are significantly closer to the ground truth. Another observation is that the planar motion (X-Y direction in the camera frame) is much more accurately estimated than height differences (Z direction in the camera frame).\par


We believe that the main cause of both of these observations is the small depth of the 3D points in the scene compared to the depth of the scene in the camera frame. This results in the existence of multiple essential matrices and homographies that can reproject the points with an error smaller than $1px$. This should be confirmed by seeding RANSAC differently (known problem in OpenCV).\par

The most accurate trajectories were generated using RoMa. RoMa produces a huge number of high-quality matches compared to other methods. Furthermore, the matches are well distributed over the pixel coordinates, resulting in good approximations of the transformations. The second most accurate trajectories were generated using SIFT+LRT. This is a surprising result, once again attesting to the incredibly well-designed SIFT features. SIFT keypoints were used directly, without any non-maxima-suppresion (NMS). The third-best trajectories were generated using MASt3R. MASt3R hugely benefits from its pre-training, being able to also reconstruct 3D scenes. Furthermore, it also produces well-distributed keypoints over the pixel coordinates. Unfortunately, MASt3R significantly down-samples the input images (to $512\times288$), which is probably the one of the reasons for its failures on our dataset.\par

SuperPoint, with SuperGlue or LightGlue produced the worst trajectories. One reason for this is the failure to match keypoints under large rotations ($>45^\circ$). The second likely reason, is the relatively small number of matches produced, sometimes resulting in matches that are not well-distributed, leading to bad estimations of the essential matrix or homography.

\FloatBarrier
\section{Conclusion}

In this paper, we presented an empirical evaluation of several image-matching techniques in the context of visual odometry for a downward-facing camera mounted on a UAV. The evaluation is done on a novel synthetic dataset that contains highly realistic aerial photography with exact ground-truth trajectories. We showed that given the 3D characteristics of the scenes, it is probably better to estimate the trajectory by fitting homographies to the matched points, rather than essential matrices. We also showed that some of the recent state-of-the-art methods do not outperform hand-crafted feature extractors on this dataset, mainly due to un-robustness to rotations. The best results were achieved using RoMa - a dense image matcher.

In the future, this work will be expanded in several directions. First, we will expand the ``naive'' approach to VO, by detecting good keyframes and filtering the matches that are used for estimating the transformations (e.g. NMS). Second, we will explore semantic segmentation of the images, to better choose appropriate matches and switch between fitting an essential matrix and homography. Lastly, we will expand the approach with registering the UAV images with satellite images, to deal with the drift that is bound to happen in a purely VO approach to localization.

\vspace{1em}
\noindent \textbf{Acknowledgements.} This work was supported ARIS research program P2-0214.




{\footnotesize
\bibliographystyle{splncs04}
\bibliography{vo-bib}

@article{SIFT,
  acmid = {996342},
  address = {Hingham, MA, USA},
  author = {Lowe, David G.},
  description = {Distinctive Image Features from Scale-Invariant Keypoints},
  issn = {0920-5691},
  journal = {Int. J. Comput. Vision},
  month = nov,
  number = 2,
  numpages = {20},
  pages = {91--110},
  publisher = {Kluwer Academic Publishers},
  title = {Distinctive Image Features from Scale-Invariant Keypoints},
  volume = 60,
  year = 2004
}

@article{SURF,
  author = {Bay, Herbert and Ess, Andreas and Tuytelaars, Tinne and Gool, Luc Van},
  description = {ScienceDirect.com - Computer Vision and Image Understanding - Speeded-Up Robust Features (SURF)},
  issn = {1077-3142},
  journal = {Computer Vision and Image Understanding},
  note = {Similarity Matching in Computer Vision and Multimedia},
  number = 3,
  pages = {346 - 359},
  title = {Speeded-Up Robust Features (SURF)},
  volume = 110,
  year = 2008
}

@inproceedings{BRIEF,
  acmid = {1888148},
  address = {Berlin, Heidelberg},
  author = {Calonder, Michael and Lepetit, Vincent and Strecha, Christoph and Fua, Pascal},
  booktitle = {Proceedings of the 11th European conference on Computer vision: Part IV},
  description = {BRIEF},
  isbn = {3-642-15560-X, 978-3-642-15560-4},
  location = {Heraklion, Crete, Greece},
  numpages = {15},
  pages = {778--792},
  publisher = {Springer-Verlag},
  series = {ECCV'10},
  title = {BRIEF: binary robust independent elementary features},
  year = 2010
}

@inproceedings{superpoint,
  author    = {Daniel DeTone and
               Tomasz Malisiewicz and
               Andrew Rabinovich},
  title     = {SuperPoint: Self-Supervised Interest Point Detection and Description},
  booktitle = {CVPR Deep Learning for Visual SLAM Workshop},
  year      = {2018},
}

@inproceedings{superglue,
  title     = {{SuperGlue}: Learning Feature Matching with Graph Neural Networks},
  author    = {Paul-Edouard Sarlin and
               Daniel DeTone and
               Tomasz Malisiewicz and
               Andrew Rabinovich},
  booktitle = {CVPR},
  year      = {2020},
}

@inproceedings{lightglue,
  author    = {Philipp Lindenberger and
               Paul-Edouard Sarlin and
               Marc Pollefeys},
  title     = {{LightGlue: Local Feature Matching at Light Speed}},
  booktitle = {ICCV},
  year      = {2023}
}

@article{roma,
title={{RoMa: Robust Dense Feature Matching}},
author={Edstedt, Johan and Sun, Qiyu and Bökman, Georg and Wadenbäck, Mårten and Felsberg, Michael},
journal={IEEE Conference on Computer Vision and Pattern Recognition},
year={2024}
}

@inproceedings{dust3r,
      title={DUSt3R: Geometric 3D Vision Made Easy}, 
      author={Shuzhe Wang and Vincent Leroy and Yohann Cabon and Boris Chidlovskii and Jerome Revaud},
      booktitle = {CVPR},
      year = {2024}
}

@INPROCEEDINGS{ORB,

  author={Rublee, Ethan and Rabaud, Vincent and Konolige, Kurt and Bradski, Gary},

  booktitle={2011 International Conference on Computer Vision}, 

  title={ORB: An efficient alternative to SIFT or SURF}, 

  year={2011},

  volume={},

  number={},

  pages={2564-2571},

  doi={10.1109/ICCV.2011.6126544}}

@INPROCEEDINGS{KITTI,
  author={Geiger, Andreas and Lenz, Philip and Urtasun, Raquel},
  booktitle={2012 IEEE Conference on Computer Vision and Pattern Recognition}, 
  title={Are we ready for autonomous driving? The KITTI vision benchmark suite}, 
  year={2012},
  volume={},
  number={},
  pages={3354-3361},
  doi={10.1109/CVPR.2012.6248074}}

@inproceedings{dedode,
  title={{DeDoDe: Detect, Don't Describe --- Describe, Don't Detect for Local Feature Matching}},
  author = {Johan Edstedt and Georg Bökman and Mårten Wadenbäck and Michael Felsberg},
  booktitle={2024 International Conference on 3D Vision (3DV)},
  year={2024},
  organization={IEEE}
}

@INPROCEEDINGS{SILK,
  author={Gleize, Pierre and Wang, Weiyao and Feiszli, Matt},
  booktitle={2023 IEEE/CVF International Conference on Computer Vision (ICCV)}, 
  title={SiLK: Simple Learned Keypoints}, 
  year={2023},
  volume={},
  number={},
  pages={22442-22451},
  doi={10.1109/ICCV51070.2023.02056}}

@article{ALIKED,
    title = {ALIKED: A Lighter Keypoint and Descriptor Extraction Network via Deformable Transformation},
    url = {https://arxiv.org/pdf/2304.03608.pdf},
    doi = {10.1109/TIM.2023.3271000},
    journal = {IEEE Transactions on Instrumentation and Measurement},
    author = {Zhao, Xiaoming and Wu, Xingming and Chen, Weihai and Chen, Peter C. Y. and Xu, Qingsong and Li, Zhengguo},
    year = {2023},
    volume = {72},
    pages = {1-16},
}

@article{LoFTR,
  title={{LoFTR}: Detector-Free Local Feature Matching with Transformers},
  author={Sun, Jiaming and Shen, Zehong and Wang, Yuang and Bao, Hujun and Zhou, Xiaowei},
  journal={CVPR},
  year={2021}
}

@article{ASpanFormer,
  title={ASpanFormer: Detector-Free Image Matching with Adaptive Span Transformer},
  author={Chen, Hongkai and Luo, Zixin and Zhou, Lei and Tian, Yurun and Zhen, Mingmin and Fang, Tian and McKinnon, David and Tsin, Yanghai and Quan, Long},
  journal={European Conference on Computer Vision (ECCV)},
  year={2022}
}

@inproceedings{CasMTR,
      title={Improving Transformer-based Image Matching by Cascaded Capturing Spatially Informative Keypoints},
      author={Cao, Chenjie and Fu, Yanwei},
      booktitle={Proceedings of the IEEE/CVF international conference on computer vision},
      year={2023}
}

@misc{mast3r,
      title={Grounding Image Matching in 3D with MASt3R}, 
      author={Vincent Leroy and Yohann Cabon and Jérôme Revaud},
      year={2024},
      eprint={2406.09756},
      archivePrefix={arXiv},
      primaryClass={cs.CV},
      url={https://arxiv.org/abs/2406.09756}, 
}

@article{mars,
author = {Johnson, Andrew E. and Cheng, Yang and Trawny, Nikolas and Montgomery, James F. and Schroeder, Steven and Chang, Johnny and Clouse, Daniel and Aaron, Seth and Mohan, Swati},
title = {Implementation of a Map Relative Localization System for Planetary Landing},
journal = {Journal of Guidance, Control, and Dynamics},
volume = {46},
number = {4},
pages = {618-637},
year = {2023},
doi = {10.2514/1.G006780},

URL = { 
    
        https://doi.org/10.2514/1.G006780
    
    

},
eprint = { 
    
        https://doi.org/10.2514/1.G006780
    
    

}
}

@InProceedings{hpatches_2017_cvpr,
author={Vassileios Balntas and Karel Lenc and Andrea Vedaldi and Krystian Mikolajczyk},
title = {HPatches: A benchmark and evaluation of handcrafted and learned local descriptors},
booktitle = {CVPR},
year = {2017}}

@ARTICLE {inloc,
author = {H. Taira and M. Okutomi and T. Sattler and M. Cimpoi and M. Pollefeys and J. Sivic and T. Pajdla and A. Torii},
journal = {IEEE Transactions on Pattern Analysis and Machine Intelligence},
title = {InLoc: Indoor Visual Localization with Dense Matching and View Synthesis},
year = {2021},
volume = {43},
number = {04},
issn = {1939-3539},
pages = {1293-1307},
doi = {10.1109/TPAMI.2019.2952114},
publisher = {IEEE Computer Society},
address = {Los Alamitos, CA, USA},
month = {apr}
}
}

\end{document}